\documentclass{article}
\usepackage{spconf,amsmath,epsfig}
\usepackage{graphicx}
\usepackage{amssymb}
\usepackage{booktabs}
\usepackage{float}
\usepackage{bm}
\usepackage{booktabs}
\usepackage{multirow}
\usepackage{makecell}
\usepackage{color}
\usepackage{algorithm}
\usepackage{algorithmicx}
\usepackage[pagebackref,breaklinks]{hyperref}

\usepackage{algpseudocode}
\usepackage{subfigure}
\let\OLDthebibliography\thebibliography
\renewcommand\thebibliography[1]{
  \OLDthebibliography{#1}
  \setlength{\parskip}{0pt}
  \setlength{\itemsep}{0pt plus 0.3ex}
}

\begin{document}\sloppy

\def\x{{\mathbf x}}
\def\L{{\cal L}}

\title{Multi-level Metric Learning for Few-shot Image Recognition}
%
\name{Haoxing Chen, Huaxiong Li, Yaohui Li, Chunlin Chen}
\address{Nanjing University\\ \small\tt \{haoxingchen, yaohuili\}@smail.nju.edu.cn, \{huaxiongli, clchen\}@nju.edu.cn}

\maketitle

\begin{abstract}
Few-shot learning is devoted to training a model on few samples. Most of these approaches learn a model based on a pixel-level or global-level feature representation. However, using global features may lose local information, and using pixel-level features may lose the contextual semantics of the image. 
Moreover, such works can only measure the relations between them on a single level, which is not comprehensive and effective. And if query images can simultaneously be well classified via three distinct level similarity metrics, the query images within a class can be more tightly distributed in a smaller feature space, generating more discriminative feature maps. Motivated by this, we propose a novel Part-level Embedding Adaptation with Graph (PEAG) method to generate task-specific features. Moreover, a Multi-level Metric Learning (MML) method is proposed, which not only calculates the pixel-level similarity but also considers the similarity of part-level features and global-level features. Extensive experiments on popular few-shot image recognition datasets prove the effectiveness of our method compared with the state-of-the-art methods. Our code is available at \url{https://github.com/chenhaoxing/M2L}.
\end{abstract}
\section{Introduction}
Humans can learn novel concepts and objects with just a few samples. Recently, many methods were proposed to learn new concepts with limited labeled data, such as semi-supervised learning~\cite{cvprWangKG0K20}, zero-shot learning~\cite{cvprYuJHZ20,cvprWuZZLZW20}, and few-shot learning~\cite{finn2017model,snell2017prototypical,sung2018learning,li2019revisiting,simon2020adaptive,chen2020multi}. Facing with the problem of data scarcity, these three paradigms propose solutions from different perspectives. Semi-supervised learning aims to train a model with few labeled data and a large amount of unlabeled data, and zero-shot learning devoted to identifying unseen categories with no labeled data, while few-shot learning focuses on learning new concepts with few labeled data. We propose a novel few-shot learning method to address the problem of data scarcity in this paper.

The few-shot learning methods can be roughly classified into two categories: meta-learning based methods~\cite{finn2017model,sun2019meta} and metric-learning based methods~\cite{snell2017prototypical,sung2018learning,li2019revisiting,simon2020adaptive,chen2020multi}. Metric-based few-shot learning methods have achieved remarkable success due to their fewer parameters and effectiveness. In this work, we focus on this branch.

The basic idea of the metric-learning based few-shot learning method is to learn a good metric to calculate the similarity between query images and the support set. Therefore, how to learn good feature embedding representation and similarity metric are the key problem of metric-learning based few-shot learning method. For feature embedding representation, Prototypical Networks~\cite{snell2017prototypical} and Relation Networks~\cite{sung2018learning} adopt image-level feature representations. However, due to the scarcity of data, it is not sufficient to measure the relation at the image-level~\cite{snell2017prototypical,sung2018learning}. Recently, CovaMNet~\cite{li2019distribution}, DN4~\cite{li2019revisiting} and MATANet~\cite{chen2020multi} introduce local representations (LRs) into few-shot learning and utilize these LRs to represent the image features, which can achieve better recognition results. 

For similarity metrics, these existing methods calculate similarities by different metrics. For example, Relation Networks~\cite{sung2018learning} proposes a network to learn the most suitable image-level similarity metric functions. DN4~\cite{chen2020multi} proposes a cosine-based image-to-class metric to measure the similarity on pixel-level.

However, global-level features lose local semantic information and pixel-level features lose contextual semantics, thus all methods mentioned above are not effective for few-shot learning.
Moreover, these methods only calculate similarities on a single level, i.e., pixel-level or image-level, which is not effective enough. Intuitively, under the few-shot learning setting, the features obtained by adopting a single similarity measure are not comprehensive, and the single similarity measure may lead to a certain similarity deviation, thus reducing the generalization ability of the model. It is necessary to adopt multi-level similarity metric, generating more discriminative features rather than using a single measure.

To this end, we propose part-level embedding adaptation with graph (PEAG) method and multi-level metric learning method (MML).  In PEAG, we divide each image into patches and get part-level features. Then, we utilize Graph Convolutional Network (GCN) to generate task-spcific features. Finally, we adpot a nearest neighbor matching module to get part-level similarity. 
In MML, in addition to component-level measures, we also use global-level measures and pixel-level measures to provide complementary information for a more compact measurement space. In MML, we also use global-level and pixel-level metrics to provide complementary information, and images within a class can be more tightly distributed in a smaller feature space.

The main contributions are summarized as follows: 
\begin{itemize}
	\setlength{\itemsep}{0pt}
	\setlength{\parsep}{0pt}
	\setlength{\parskip}{0pt}
	\item We propose a novel part-level embedding adatation with graph method, which can generate task-specific part-level features and capture the part-level semantic similarity between query images and support images. 
	\item We propose a novel multi-level metric learning method by computing the semantic similarities on pixel-level, part-level, and global-level simultaneously, aiming to find more comprehensive semantic similarities. 
	\item We conduct sufficient experiments on popular benchmark datasets to verify the advancement of our model and the performance of our model achieves the state-of-the-art.
\end{itemize}
\begin{figure*}[t]
	\centering
	\includegraphics[height=8cm,width=17cm]{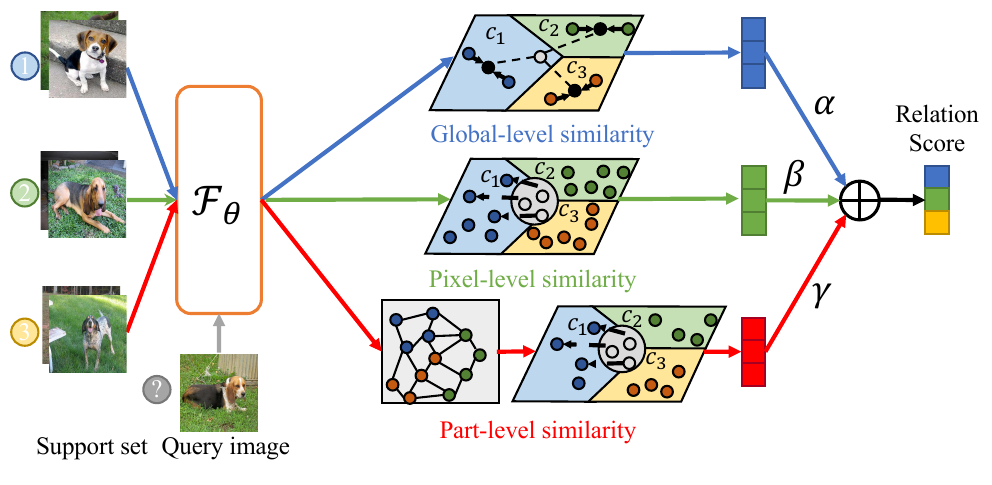}
	\caption{The framework of MML under the 3-way 2-shot image classification setting. (Best view in color.)}
	\label{mml_architecture}
\end{figure*}

\section{Related Works}
In this section, we focus on related works on metric-learning based few-shot learning model.

\subsection{Learning feature embedding representation}
Koch et al. \cite{koch2015siamese} used a Siamese Neural Network to tackle the one-shot learning problem, in which the feature extractor is of VGG styled structure and $L_1$ distance is used to measure the similarity between query images and support images.
Snell et al. \cite{snell2017prototypical} proposed Prototypical Networks, in which the Euclidean distance is used to compute the distance between class-specific prototypes. 
Li et al. \cite{li2019revisiting} proposed an image-to-class mechanism to find the relation at pixel-level, in which the image features are represented as a local descriptor collection.

\subsection{Learning similarity metric}
Sung et al.~\cite{sung2018learning} replaced the existing metric with the Relation Network, which measures the similarity between each query instance and support classes. Li et al.~\cite{li2019revisiting} proposed a Deep Nearest Neighbor Neural Network (DN4) to learn an image-to-class metric by measuring the cosine similarity between the deep local descriptors of a query instance and its neighbors from each support class. Li et al.~\cite{li2019distribution}explored the distribution consistency-based metric by introducing local covariance representation and deep covariance metric. Unlike these methods, the proposed MML measures the similarity at three different feature levels, i.e., pixel-level, part-level, and distribution-level.

\section{Problem Definition and Formulation}
Standard few-shot image recognition problems are often formalized as \emph{N}-way \emph{M}-shot classification problem, in which models are given \emph{M} seen images from each of \emph{N} classes, and required to correctly classify unseen images. 
Different from traditional image recognition tasks, few-shot learning aims to classify novel classes after training. This requires
that samples used for training, validation, and testing should
come from disjoint label space.
To be more specific, given a dataset of visual concepts $\mathcal{C}$, we devide it into three parts: $\mathcal{C}_{train}$, $\mathcal{C}_{val}$ and $\mathcal{C}_{test}$, and their label space satisfy $\mathcal{L}_{train}\cap\mathcal{L}_{val}\cap\mathcal{L}_{test}=\emptyset$.

To obtain a trained model, we train our model in an episodic way. That is, in each episode, a new task is randomly sampled from the training set $\mathcal{C}_{train}$ to train the current model. Each task consists of two subsets, including support set $\mathcal{A}_{\mathcal{S}}$ and query set $\mathcal{A}_{\mathcal{Q}}$. The $\mathcal{A}_{\mathcal{S}}$ contains $\mathcal{N}$ previously unseen classes, with $\mathcal{M}$ samples for each class. We focus on training our model to correctly determine which category each image in the $\mathcal{A}_{\mathcal{Q}}$ belongs to. Similarly, we randomly sample tasks from $\mathcal{C}_{val}$ and $\mathcal{C}_{test}$ for meta-validation and meta-testing scenarios.

\section{Multi-level Metric Learning}
As shown in Figure 2, our MML is mainly composed of two modules: a feature extractor $\mathcal{F}_{\theta}$, a multi-level metric-learning module. All images are first fed into the $\mathcal{F}_{\theta}$ to get feature embeddings. Then, the multi-level metric-learning module calculates similarities on part-level, pixel-level, and distribution-level simultaneously. Finally, we fuse these three similarities together. All the modules can be trained jointly in an end-to-end manner.
\subsection{Part-level Metric}
We divide each image into $H\times W$ patches evenly, and input each patch into $\mathcal{F}_{\theta}$ separately to generate part-level descriptors. In order to further enhance the representation ability of features, we adopt a pyramid structure. Thus, under \emph{N-way M-shot} few-shot learning setting, given a query image $q$ and support class $\mathcal{S}_n, n=\{1, ..., N\}$, through $\mathcal{F}_{\theta}$ and global-average pooling (GAP) layer, we can get the part-level descriptors:
$\mathcal{X}^{\rm part}_q\in\mathbb{R}^{C\times K}$ and $\mathcal{X}^{\rm part}_{\mathcal{S}_n}\in\mathbb{R}^{C\times MK}$. Specifically, we adopt two image patch division strategies of size 2$\times$2 and 3$\times$3 to obtain 13 part-level descriptors.  

To generate task-specific support features, we propose a novel part-level embedding adaptation with graph (PEAG) method. Specifically, we concatenate all support features and get $\mathcal{X}^{\rm part}_{\mathcal{S}}\in\mathbb{R}^{C\times NMK}$. Then, we construct the degree matrix $A\in\mathbb{R}^{NMK\times NMK}$ to represent the similarity between patches in the support set.
If two patches $p_i$ and $p_j$ have the same semantics, then we set the corresponding element $A_{ij}$ to 1, otherwise to 0. Based on $A$, we build the adjacency matrix $S$:
\begin{gather}
	S=D^{-\frac{1}{2}}(A+I)D^{-\frac{1}{2}}
\end{gather}
where $I\in\mathbb{R}^{NMK\times NMK}$ is the identity matrix and $D$ is the diagonal matrix ($D_{ii}=\sum_{j}A_{ij}+1$). Let $\varPsi^0=\mathcal{X}^{\rm part}_{\mathcal{S}}$, the relationship between patches could be propagated based on $S$:
\begin{gather}
	\varPsi^{t+1}={\rm ReLU}(S\varPsi^{t}W), t=\{0,...,T-1\}
\end{gather}
where $W$ is a learned feature transformation matrix. After propagate the embedding set $T$ times, we can get the final propagated embedding set $\varPsi^{T} = \mathcal{X}^{\rm part}_{\mathcal{S}'}$.

Then, for each class $\mathcal{X}^{\rm part}_{\mathcal{S}'_n}$, we calculate the correlation matrix $R^{\rm part}\in\mathbb{R}^{K\times MK}$ between the query image and the support class $n$ on part-level:
\begin{gather}
	\mathcal{R}^{\rm part} = \frac{(\mathcal{X}^{\rm part}_q)^\top\mathcal{X}^{\rm part}_{\mathcal{S}'_n}}
	{
		\left\|\mathcal{X}^{\rm part}_q\right\| \cdot  \left\|\mathcal{X}^{\rm part}_{\mathcal{S}'_n}\right\|} 
\end{gather}
$\mathcal{R}^{\rm part}_{i,j}$ is $(i,j)$ element of $\mathcal{R}^{\rm part}$ reflecting the distance between the $i$-th descriptors\ of the query image and the $j$-th descriptor of support clss $n$. Each row in $\mathcal{R}^{\rm part}$ represents the semantic relation of each descriptor in the query image to all descriptors of all images in the support class.
For each patch of the query image $q$, we find its most similar descriptor. Then, we sum $K$ selected part-level descriptors as the part-level similarity between the query image and the support class $n$:
\begin{equation}
	\mathcal{D}_{\rm part}(q, \mathcal{S}_n)=\sum_{i=1}^{K}{\rm Top1}(\mathcal{R}_{i}^{\rm part} )
\end{equation}
where ${\rm Top}(\cdot)$ means selecting the largest elements in each row of the $R^{\rm part}$.

\subsection{Pixel-level Metric}
Following~\cite{li2019distribution,li2019revisiting,chen2020multi}, given a query image $q$ and a certain support class $\mathcal{S}_n$, through feature extractor $\mathcal{F}_{\theta}$, we can get the feature representation $\mathcal{F}_{\theta}(q)\in\mathbb{R}^{C\times H\times W}$ and $\mathcal{F}_{\theta}(\mathcal{S}_n)\in\mathbb{R}^{M\times C\times H\times W}$, respectively. The $\mathcal{F}_{\theta}(q)$ can be regard as a set of $H\times W$ $C$-dimensional LRs:
\begin{gather}
	\mathcal{L}^{\rm pixel}_{q}=[u^{\rm pixel}_1,...,u^{\rm pixel}_{HW}]\in\mathbb{R}^{C\times HW} 
\end{gather}
Also, the $\mathcal{F}_{\theta}(\mathcal{S}_n)$ can be regards as 
\begin{gather}	
	\mathcal{L}^{\rm pixel}_{\mathcal{S}_n}=[v^{\rm pixel}_1,...,v^{\rm pixel}_{MHW}]\in\mathbb{R}^{C\times MHW} 
\end{gather}
Then, we calculate the correlation matrix $R^{\rm pixel}\in\mathbb{R}^{HW\times MHW}$  between the query image and the support class on pixel-level and select the largest element in each row of the correlation matrix:
\begin{gather}
	\mathcal{R}^{\rm pixel} = \frac{(u^{\rm pixel}_i)^\top v^{\rm pixel}_j}
	{
		\left\|u^{\rm pixel}_i \right\| \cdot  \left\|v^{\rm pixel}_j\right\|} 	
	\\
	\mathcal{D}_{\rm pixel}(q, \mathcal{S}_n)= \sum_{i=1}^{HW} {\rm Top1}(\mathcal{R}_{i}^{\rm pixel} )
\end{gather}

\subsection{Global-level Metric}
We adpot Prototypical Networks~\cite{snell2017prototypical} as our global-level similarity metric. Prototypical Networks computes the empirical mean of global convulution embeddings as the prototype representation of each category $n$:
\begin{equation}
	c_n = \frac{1}{M} \sum_{M}^{i=1} {\rm GAP}(\mathcal{F}_{\theta}(\mathcal{S}_n^i))
\end{equation}
where $p_n\in\mathbb{R}^{K}$. Similarly, given a query image $Q$, we can get its global convulution embeddings $\mathcal{X}^{\rm global}_Q\in\mathbb{R}^{K}$.
Then, Prototypical Networks utilized Euclidean distance as the distance metric and assigns a probability over class $n$:
\begin{equation}
	\mathcal{D}_{\rm global}(q, \mathcal{S}_n)= -d(\mathcal{X}^{\rm global}_Q, c_n)
\end{equation}

\begin{table}[t]
	\centering
	\begin{tabular}{ccccc}
		\toprule
		\textbf{Dataset} &\textbf{$N_{all}$}&\textbf{$N_{train}$}&\textbf{$N_{val}$}&\textbf{$N_{test}$}
		\\
		\midrule
		\emph{mini}ImageNet&100&64&16&20\\
		\emph{tiered}ImageNet&608&351&97&160\\
		CIFAR-100&100&64&16&20\\		
		FC100&100&60&20&20\\
		\bottomrule
	\end{tabular}	
	\caption{The splits of evaluation datasets. $N_{all}$ is the number of all classes. $N_{train}$, $N_{val}$ and $N_{test}$ indicate the number of classes in training set, validation set and test set.}
\end{table}

\subsection{Fusion Layer}
Since three different level similarities have been calculated, we need to design a fusion module to integrate them. Specifically, the final similarity and probability over any class $n$ can be obtained by the following equation: 
\begin{gather}
	P_{part}(y=n|q) =  \frac{\mathcal{D}_{part}(q, \mathcal{S}_n)}{\sum_{i=1}^{N}\mathcal{D}_{part}(q, \mathcal{S}_n)}\\
	P_{pixel}(y=n|q) =\frac{\mathcal{D}_{pixel}(q, \mathcal{S}_n)}{\sum_{i=1}^{N}\mathcal{D}_{pixel}(q, \mathcal{S}_n)}\\
	P_{global}(y=n|q) =\frac{\mathcal{D}_{global}(q, \mathcal{S}_n)}{\sum_{i=1}^{N}\mathcal{D}_{global}(q, \mathcal{S}_n)}\\
	\begin{aligned}
		P(y=n|q) =& \alpha P_{part}(y=n|q) + 	\beta P_{global}(y=n|q)\\&  +\gamma 	P_{pixel}(y=n|q)
	\end{aligned}
\end{gather}
where $y$ is the label of $q$, $\alpha$, $\beta$ and $\gamma$ are superparameters. If $y=n'$, then we can define the loss function as follows:
\begin{equation}
	\begin{aligned}
		\mathcal{L} =& -\alpha log(p_{part}(y=n'|q))-\beta log(p_{pixel}(y=n'|q))\\&-\gamma log(p_{global}(y=n'|q))
	\end{aligned}
\end{equation}

\section{Experiments}
In this section, we perform extensive experiments to verify the advance and effectiveness of MML. 
\subsection{Datasets}
To verify the advance and effectiveness of our proposed MML, we performed experiments on four benchmark datasets.

\textbf{ImageNet derivatives:} Both \emph{mini}ImageNet~\cite{vinyals2016matching} dataset and \emph{tiered}ImageNet~\cite{ren18iclr} dataset are subsets of ImageNet \cite{deng2009imagenet}. The \emph{mini}ImageNet dataset consists 100 classes, each of which contains 600 samples, and the \emph{tiered}ImageNet contains 608 classes.

\textbf{CIFAR derivatives:} 
Both CIFAR-FS~\cite{r2d2} dataset and FC100~\cite{tadam} dataset are subsets of CIFAR-100. Both of them consist 100 classes. 

The partition of all data sets is shown in Table 1. All images are resized to $84\times 84$. 

\begin{table}[t]
	\centering
	\begin{tabular}{ccccc}
		\toprule
		\multirow{3}{*}{\textbf{$\alpha$}}
		&\multirow{3}{*}{\textbf{$\beta$}} &\multirow{3}{*}{\textbf{$\gamma$}}& \multicolumn{2}{c}{\textbf{5-Way Accuracy($\%$)}}
		\\
		& & & 1-shot & 5-shot \\
		\midrule
		1&0&0&\textbf{67.29$\pm$\footnotesize{0.23}} &78.49$\pm$\footnotesize{0.21}\\
		0&1&0&64.12$\pm$\footnotesize{0.23}  & 78.55$\pm$\footnotesize{0.21}   \\
		0&0&1& 61.86$\pm$\footnotesize{0.24}   &  79.03$\pm$\footnotesize{0.21}  \\
		1&1&0& 66.85$\pm$\footnotesize{0.23}   &  80.52$\pm$\footnotesize{0.20}  \\   
		1&0&1& 61.65$\pm$\footnotesize{0.24}   &  78.87$\pm$\footnotesize{0.21}  \\   
		0&1&1& 61.95$\pm$\footnotesize{0.24}   &  78.85$\pm$\footnotesize{0.21}  \\   
		1&1&1& 64.77$\pm$\footnotesize{0.23}   &  79.93$\pm$\footnotesize{0.20}  \\   
		1&0.5&0.5& 66.72$\pm$\footnotesize{0.23}  &\textbf{81.01$\pm$\footnotesize{0.20} } \\ 
		1&0.1&0.1& \textbf{67.58$\pm$\footnotesize{0.23}} &\textbf{81.41$\pm$\footnotesize{0.20}} \\ 
		\bottomrule
	\end{tabular}
	\caption{Ablation study on \emph{mini}ImageNet. (Top two performances are in bold font.)}
\end{table}

\begin{table*}[t]
	\centering
	\begin{tabular}{cccccc}
		\toprule
		\label{imagenet}
		\multirow{2}{*}{\textbf{Model}} &\multirow{2}{*}{\textbf{Backbone}} 
		&\multicolumn{2}{c}{\textbf{\emph{mini}ImageNet}} &\multicolumn{2}{c}{\textbf{\emph{tiered}ImageNet}} \\
		\cmidrule{5-6}		\cmidrule{3-4}
		& & 1-shot & 5-shot & 1-shot & 5-shot\\
		\midrule
		Prototypical Networks \cite{snell2017prototypical}&Conv-64F& 49.42$\pm$\footnotesize{0.78} & 68.20$\pm$\footnotesize{0.66}&53.31$\pm$\footnotesize{0.89}&72.69$\pm$\footnotesize{0.74} \\
		Relation Networks \cite{sung2018learning}&Conv-64F& 50.44 $\pm$\footnotesize{0.82} & 65.32$\pm$\footnotesize{0.77}&54.48$\pm$\footnotesize{0.93}&71.32$\pm$\footnotesize{0.78}\\
		DN4 \cite{li2019revisiting}&Conv-64F& 51.24 $\pm$\footnotesize{0.74} & 71.02$\pm$\footnotesize{0.64}&53.37$\pm$\footnotesize{0.86}&74.45$\pm$\footnotesize{0.70}\\
		Prototypical Networks \cite{snell2017prototypical}&ResNet-12&62.59$\pm$\footnotesize{0.85} & 78.60$\pm$\footnotesize{0.16} &68.37$\pm$\footnotesize{0.23} & 83.43$\pm$\footnotesize{0.16} \\
		TADAM \cite{tadam}&ResNet-12& 58.50$\pm$\footnotesize{0.30}&76.70$\pm$\footnotesize{0.30}& -  &-\\
		MeaOptNet \cite{cvprLeeMRS19}& ResNet-12&62.64$\pm$\footnotesize{0.61} & 78.63$\pm$\footnotesize{0.46}& 65.99$\pm$\footnotesize{0.72}  & 81.56$\pm$\footnotesize{0.53}  \\
		DSN-MR \cite{simon2020adaptive}& ResNet-12&
		64.60$\pm$\footnotesize{0.72} & 79.51$\pm$\footnotesize{0.50}&
		67.39$\pm$\footnotesize{0.82} & 82.85$\pm$\footnotesize{0.56}  \\
		FEAT \cite{feat} & ResNet12 & 66.78$\pm$\footnotesize{0.20}& \textbf{82.05$\pm$\footnotesize{0.14}}& 67.39$\pm$\footnotesize{0.82}& 82.85$\pm$\footnotesize{0.56} \\
		GLoFA \cite{glofa} & ResNet12 &  66.12$\pm$\footnotesize{0.42}& 81.37$\pm$\footnotesize{0.33}&\textbf{69.75$\pm$\footnotesize{0.33}}& 83.58$\pm$\footnotesize{0.42}  \\
		Fine-tuning \cite{finn2017model}&WRN-28-10& 57.73$\pm$\footnotesize{0.62} & 78.17$\pm$\footnotesize{0.49}& 66.58$\pm$\footnotesize{0.70}& \textbf{85.55$\pm$\footnotesize{0.48}} \\
		AWGIM~\cite{awg} &WRN-28-10& 63.12$\pm$\footnotesize{0.08} & 78.40$\pm$\footnotesize{0.11}& 67.69$\pm$\footnotesize{0.11}& 82.82$\pm$\footnotesize{0.13} \\		
		\midrule
		\textbf{PEAG} &ResNet-12& \textbf{67.29$\pm$\footnotesize{0.23}} &78.49$\pm$\footnotesize{0.21}&68.89$\pm$\footnotesize{0.25}&82.08$\pm$\footnotesize{0.21}\\
		\textbf{MML}  &ResNet-12& \textbf{67.58$\pm$\footnotesize{0.23}} &\textbf{81.41$\pm$\footnotesize{0.20}}&\textbf{71.38$\pm$\footnotesize{0.25}} & \textbf{84.65$\pm$\footnotesize{0.20}}\\
		\bottomrule
	\end{tabular}
	\caption{Comparison with other state-of-the-art methods with $95\%$ confidence intervals on \emph{mini}ImageNet and \emph{tiered}ImageNet. (Top two performances are in bold font.)}
\end{table*}

\begin{table*}[htbp]
	\centering
	\begin{tabular}{cccccc}
		\toprule
		\label{cifar}
		\multirow{2}{*}{\textbf{Model}} &\multirow{2}{*}{\textbf{Backbone}} 
		&\multicolumn{2}{c}{\textbf{CIFAR-FS}} &\multicolumn{2}{c}{\textbf{FC100}} \\
		\cmidrule{5-6}		\cmidrule{3-4}
		& & 1-shot & 5-shot & 1-shot & 5-shot\\
		\midrule
		Prototypical Networks \cite{snell2017prototypical}&Conv-64F& 55.50$\pm$\footnotesize{0.70} & 72.00$\pm$\footnotesize{0.60}&35.30$\pm$\footnotesize{0.60}&48.60$\pm$\footnotesize{0.60} \\
		Relation Networks \cite{sung2018learning}&Conv-256F& 55.00$\pm$\footnotesize{1.00} & 69.30$\pm$\footnotesize{0.80}&-&-\\
		R2D2 \cite{r2d2}&Conv-512F& 65.30$\pm$\footnotesize{0.20} & 79.40$\pm$\footnotesize{0.10}&-&-\\
		Prototypical Networks \cite{snell2017prototypical}&ResNet-12&72.20$\pm$\footnotesize{0.70} & 83.50$\pm$\footnotesize{0.50}& 37.50$\pm$\footnotesize{0.60} & 52.50$\pm$\footnotesize{0.60} \\
		TADAM \cite{tadam}&ResNet-12& -&-& 40.10$\pm$\footnotesize{0.40}  & 56.10$\pm$\footnotesize{0.40}  \\
		MeaOptNet \cite{cvprLeeMRS19}& ResNet-12&72.60$\pm$\footnotesize{0.70} & 84.30$\pm$\footnotesize{0.50}& 41.10$\pm$\footnotesize{0.60}  & 55.50$\pm$\footnotesize{0.60}  \\
		MABAS \cite{mabas}& ResNet-12&
		73.51$\pm$\footnotesize{0.92} & 85.49$\pm$\footnotesize{0.68}&
		42.31$\pm$\footnotesize{0.75} & 57.56$\pm$\footnotesize{0.78}  \\
		Fine-tuning \cite{dhillon2019baseline}&WRN-28-10& \textbf{76.58$\pm$\footnotesize{0.68}} & \textbf{85.79$\pm$\footnotesize{0.50}}& 43.16$\pm$\footnotesize{0.59} & \textbf{57.57 $\pm$\footnotesize{0.55}} \\														
		\midrule
		\textbf{PEAG}  &ResNet-12&74.27$\pm$\footnotesize{0.23} &83.89$\pm$\footnotesize{0.20}&\textbf{43.99$\pm$\footnotesize{0.21}} & 56.47$\pm$\footnotesize{0.24}\\
		\textbf{MML}  &ResNet-12&\textbf{75.28$\pm$\footnotesize{0.23}} &\textbf{85.95$\pm$\footnotesize{0.19}}&\textbf{44.43$\pm$\footnotesize{0.21}} & \textbf{59.56$\pm$\footnotesize{0.25}}\\
		\bottomrule
	\end{tabular}
	\caption{Experimental results compared with other methods on CIFAR-FS and FC100. (Top two performances are in bold font.)}
\end{table*}

\subsection{Implementation Details}
In order to make a fair comparison with other works, we adopt the \emph{ResNet-12} network~\cite{cvprLeeMRS19} as our feature extrator $\mathcal{F}_{\theta}$. 

\emph{ResNet-12} has four residual blocks, each residual block has 3 convolutional layers with 3×3 kernel, and a $2\times 2$ max-pooling layer is added in the first residual block.

We conduct our experiments on a series of \emph{N}-way \emph{M}-shot tasks, i.e., 5-way 1-shot and 5-way 5-shot.
Following~\cite{feat}, we first pre-trained $\mathcal{F}_{\theta}$ with an MLP consisting of a single hidden layer. Then we meta-train the whole model by momentum SGD for 40 epochs.
In each epoch, we randomly sampled 200 tasks. Our batch size is set to 4, the initial learning rate is $5\times10^{-4}$, and multiplied by 0.5 every 10 epochs. During the test stage, we report the average accuracy as well as the corresponding 95\% confidence interval over these 10,000 tasks.

\subsection{Ablation Study}
To explore the effect of the multi-level metric learning module, we prune any of three similarity branches in the multi-level metric-learning module. Specifically, we change the values of $\alpha$, $\beta$ and $\gamma$, and experiment on the \emph{mini}ImageNet. 

As seen in Table 2, each part of the MML is indispensable. It can be observed that the accuracy of few-shot image recognition using only one level of features is very low. The results were significantly improved when two or three levels of features were used together, and the results were best when all three levels were used together. Specifically, compared with the method that only using pixel-level features, our MML gains 5.4\% and 3.7\% improvements.
Note that it is important to choose the appropriate hyperparameters. When $\beta$ and $\gamma$ become larger, the accuracy of the model will become worse. Appropriate $\beta$ and $\gamma$ can provide useful auxiliary information for part-level metric.

\subsection{Comparison Against Related Approaches}
\textbf{Results on ImageNet derivatives.} 
As seen from Table 3, our MML achieves the highest accuracy on \emph{mini}ImageNet with 67.58\% and 81.41\% on 5-way 1-shot and 5-way 5-shot tasks respectively, which make a great improvement compared to the previous single level metric-learning based methods. For example, our MML is 4.6\% and 8.0\% better than DSN-MR~\cite{simon2020adaptive} and Prototypical Networks~\cite{snell2017prototypical} on the 5-way 1-shot task, respectively. And our MML achieves 71.38\% and 84.65\% on \emph{tiered}ImageNet under 5-way 1-shot and 5-way 5-shot few-shot learning setting respectively, which achieves competitive performance.

\textbf{Results on CIFAR derivatives.}
Table 4 evaluates our method on two CIFAR derivatives, i.e., CIFAR-FS and FC100. It can be seen that the proposed MML obtains significant improvements compared with previous state-of-the-art methods. Specifically, compared with global-level metric-learning based methods (i.e., Relation Networks~\cite{sung2018learning}, Prototypical Networks~\cite{snell2017prototypical} and Fine-tuning \cite{dhillon2019baseline}), MML is 20.3\% and 3.5\% better than the best one of them on CIFAR-FS and FC100 under 5-way 5-shot setting.

Moreover, we can also see that the proposed PEAG achieved competitive results. For example, our PEAG is 0.8\% and 1.8\% better than FEAT~\cite{feat} and GLoFA~\cite{glofa} on \emph{mini}ImageNet under the 5-way 1-shot setting, respectively.

The reason why our MML can achieve these state-of-the-art performances is that MML can measure the semantic similarities on multiple levels, i.e., part-level, pixel-level, and global-level.

\section{Conclusion}
In this paper, we revisit the metric-learning based method and proposed novel Part-level Embedding Adaptation with Graph (PEAG) method and Multi-level Metric Learning (MML) method for few-shot image recognition, aiming to capture more comprehensive semantic similarities. Specifically, PEAG can generate task-specific part-level features and capture the part-level semantic similarity between query images and support images, and MML can measure the semantic similarities on multiple levels and produce more discriminative features. Extensive experiments show the effectiveness and the superiority of both PEAG and MML.

\bibliographystyle{IEEEbib}
\bibliography{MML}

\begin{thebibliography}{10}

\bibitem{cvprWangKG0K20}
Yaxing Wang, Salman Khan, Abel Gonzalez{-}Garcia, Joost van~de Weijer, and
  Fahad~Shahbaz Khan,
\newblock ``Semi-supervised learning for few-shot image-to-image translation,''
\newblock in {\em CVPR}, 2020, pp. 4452--4461.

\bibitem{cvprYuJHZ20}
Yunlong Yu, Zhong Ji, Jungong Han, and Zhongfei Zhang,
\newblock ``Episode-based prototype generating network for zero-shot
  learning,''
\newblock in {\em CVPR}, 2020, pp. 14032--14041.

\bibitem{cvprWuZZLZW20}
Jiamin Wu, Tianzhu Zhang, Zheng{-}Jun Zha, Jiebo Luo, Yongdong Zhang, and Feng
  Wu,
\newblock ``Self-supervised domain-aware generative network for generalized
  zero-shot learning,''
\newblock in {\em CVPR}, 2020, pp. 12764--12773.

\bibitem{finn2017model}
Chelsea Finn, Pieter Abbeel, and Sergey Levine,
\newblock ``Model-agnostic meta-learning for fast adaptation of deep
  networks,''
\newblock in {\em ICML}, 2017, vol.~70, pp. 1126--1135.

\bibitem{snell2017prototypical}
Jake Snell, Kevin Swersky, and Richard~S. Zemel,
\newblock ``Prototypical networks for few-shot learning,''
\newblock in {\em NeurIPS}, 2017, pp. 4077--4087.

\bibitem{sung2018learning}
Flood Sung, Yongxin Yang, Li~Zhang, Tao Xiang, Philip H.~S. Torr, and
  Timothy~M. Hospedales,
\newblock ``Learning to compare: Relation network for few-shot learning,''
\newblock in {\em CVPR}, 2018, pp. 1199--1208.

\bibitem{li2019revisiting}
Wenbin Li, Lei Wang, Jinglin Xu, Jing Huo, Yang Gao, and Jiebo Luo,
\newblock ``Revisiting local descriptor based image-to-class measure for
  few-shot learning,''
\newblock in {\em CVPR}, 2019, pp. 7260--7268.

\bibitem{simon2020adaptive}
Christian Simon, Piotr Koniusz, Richard Nock, and Mehrtash Harandi,
\newblock ``Adaptive subspaces for few-shot learning,''
\newblock in {\em CVPR}, 2020, pp. 4135--4144.

\bibitem{chen2020multi}
Haoxing Chen, Huaxiong Li, Yaohui Li, and Chunlin Chen,
\newblock ``Multi-scale adaptive task attention network for few-shot
  learning,''
\newblock {\em arXiv preprint arXiv:2011.14479}, 2020.

\bibitem{sun2019meta}
Qianru Sun, Yaoyao Liu, Tat{-}Seng Chua, and Bernt Schiele,
\newblock ``Meta-transfer learning for few-shot learning,''
\newblock in {\em CVPR}, 2019, pp. 403--412.

\bibitem{li2019distribution}
Wenbin Li, Jinglin Xu, Jing Huo, Lei Wang, Yang Gao, and Jiebo Luo,
\newblock ``Distribution consistency based covariance metric networks for
  few-shot learning,''
\newblock in {\em AAAI}, 2019, pp. 8642--8649.

\bibitem{koch2015siamese}
Gregory Koch, Richard Zemel, and Ruslan Salakhutdinov,
\newblock ``Siamese neural networks for one-shot image recognition,''
\newblock in {\em ICML Workshops}, 2015, vol.~2.

\bibitem{vinyals2016matching}
Oriol Vinyals, Charles Blundell, Tim Lillicrap, Koray Kavukcuoglu, and Daan
  Wierstra,
\newblock ``Matching networks for one shot learning,''
\newblock in {\em NeurIPS}, 2016, pp. 3630--3638.

\bibitem{ren18iclr}
Mengye Ren, Eleni Triantafillou, Sachin Ravi, Jake Snell, Kevin Swersky,
  Joshua~B. Tenenbaum, Hugo Larochelle, and Richard~S. Zemel,
\newblock ``Meta-learning for semi-supervised few-shot classification,''
\newblock in {\em ICLR}, 2018.

\bibitem{deng2009imagenet}
Jia Deng, Wei Dong, Richard Socher, Li{-}Jia Li, Kai Li, and Fei{-}Fei Li,
\newblock ``Imagenet: {A} large-scale hierarchical image database,''
\newblock in {\em CVPR}, 2009, pp. 248--255.

\bibitem{r2d2}
Luca Bertinetto, Jo{\~{a}}o~F. Henriques, Philip H.~S. Torr, and Andrea
  Vedaldi,
\newblock ``Meta-learning with differentiable closed-form solvers,''
\newblock in {\em ICLR}, 2019.

\bibitem{tadam}
Boris~N. Oreshkin, Pau~Rodr{\'{\i}}guez L{\'{o}}pez, and Alexandre Lacoste,
\newblock ``{TADAM:} task dependent adaptive metric for improved few-shot
  learning,''
\newblock in {\em NeurIPS}, 2018, pp. 719--729.

\bibitem{cvprLeeMRS19}
Kwonjoon Lee, Subhransu Maji, Avinash Ravichandran, and Stefano Soatto,
\newblock ``Meta-learning with differentiable convex optimization,''
\newblock in {\em CVPR}, 2019, pp. 10657--10665.

\bibitem{feat}
Han-Jia Ye, Hexiang Hu, De-Chuan Zhan, and Fei Sha,
\newblock ``Few-shot learning via embedding adaptation with set-to-set
  functions,''
\newblock in {\em CVPR}, 2020, pp. 8808--8817.

\bibitem{glofa}
Su~Lu, Han{-}Jia Ye, and De{-}Chuan Zhan,
\newblock ``Tailoring embedding function to heterogeneous few-shot tasks by
  global and local feature adaptors,''
\newblock in {\em AAAI}, 2021, pp. 8776--8783.

\bibitem{awg}
Yiluan Guo and Ngai{-}Man Cheung,
\newblock ``Attentive weights generation for few shot learning via information
  maximization,''
\newblock in {\em CVPR}, 2020, pp. 13496--13505.

\bibitem{mabas}
Jaekyeom Kim, Hyoungseok Kim, and Gunhee Kim,
\newblock ``Model-agnostic boundary-adversarial sampling for test-time
  generalization in few-shot learning,''
\newblock in {\em ECCV}, 2020, vol. 12346, pp. 599--617.

\bibitem{dhillon2019baseline}
Guneet~Singh Dhillon, Pratik Chaudhari, Avinash Ravichandran, and Stefano
  Soatto,
\newblock ``A baseline for few-shot image classification,''
\newblock in {\em ICLR}, 2020.

\end{thebibliography}

\end{document}